\RequirePackage{fix-cm}
\documentclass[smallextended]{svjour3}       
\smartqed  
\usepackage{graphicx}
\usepackage[utf8]{inputenc}
\usepackage[T1]{fontenc}

\usepackage{amsmath,amsthm} 
\usepackage{graphicx}  
\usepackage{amssymb}
\usepackage{algorithm}
\usepackage[noend]{algpseudocode}
\usepackage{multirow}
\usepackage{caption}
\usepackage{subcaption}
\usepackage{array}
\usepackage{rotating} 

\usepackage{bm}
\usepackage[switch]{lineno}
\begin{document}

\title{Federated Learning for Healthcare Informatics
}


\author{Jie Xu \and Benjamin S. Glicksberg \and Chang Su \and Peter Walker \and Jiang Bian \and Fei Wang$^*$ 
}


\institute{J. Xu, C. Su and F. Wang$^*$ \at
              Department of Healthcare Policy and Research, Weill Cornell Medicine, New York, New York, USA \\
              \email{few2001@med.cornell.edu}           
           \and
           B. S. Glicksberg \at
              Institute for Digital Health, Icahn School of Medicine at Mount Sinai, New York, New York, USA
            \and
        P. Walker \at
        U.S. Department of Defense Joint Artificial Intelligence Center, Washington, D.C., USA
        \and
        J. Bian \at Department of Health Outcomes and Biomedical Informatics, College of Medicine. University of Florida, Gainesville, Florida, USA.
}

\date{Received: date / Accepted: date}

\maketitle

\begin{abstract}
With the rapid development of computer software and hardware technologies, more and more healthcare data are becoming readily available from clinical institutions, patients, insurance companies and pharmaceutical industries, among others. This access provides an unprecedented opportunity for data science technologies to derive data-driven insights and improve the quality of care delivery. Healthcare data, however, are usually fragmented and private making it difficult to generate robust results across populations. For example, different hospitals own the electronic health records (EHR) of different patient populations and these records are difficult to share across hospitals because of their sensitive nature. This creates a big barrier for developing effective analytical approaches that are generalizable, which need diverse, ``big data". Federated learning, a mechanism of training a shared global model with a central server while keeping all the sensitive data in local institutions where the data belong, provides great promise to connect the fragmented healthcare data sources with privacy-preservation. The goal of this survey is to provide a review for federated learning technologies, particularly within the biomedical space. In particular, we summarize the general solutions to the statistical challenges, system challenges and privacy issues in federated learning, and point out the implications and potentials in healthcare.
\end{abstract}

\section{Introduction}
The recent years have witnessed a surge of interest related to healthcare data analytics, due to the fact that more and more such data are becoming readily available from various sources including clinical institutions, patient individuals, insurance companies and pharmaceutical industries, among others. This provides an unprecedented opportunity for the development of computational techniques to dig data-driven insights for improving the quality of care delivery \cite{wang2019ai,miotto2018deep}.

Healthcare data are typically fragmented because of the complicated nature of the healthcare system and processes. For example, different hospitals may be able to access the clinical records of their own patient populations only. These records are highly sensitive with protected health information (PHI) of individuals. Rigorous regulations, such as the Health Insurance Portability and Accountability Act (HIPAA) \cite{gostin2001national}, have been developed to regulate the process of accessing and analyzing such data. This creates a big challenge for modern data mining and machine learning (ML) technologies, such as deep learning \cite{lecun2015deep}, which typically requires a large amount of training data.
 
Federated learning is a paradigm with a recent surge in popularity as it holds great promise on learning with fragmented sensitive data. Instead of aggregating data from different places all together, or relying on the traditional discovery then replication design, it enables training a shared global model with a central server while keeping the data in local institutions where the they originate.

The term ``federated learning" is not new. In 1976, Patrick Hill, a philosophy professor, first developed the Federated Learning Community (FLC) to bring people together to jointly learn, which helped students overcome the anonymity and isolation in large research universities~\cite{hill1985rationale}. Subsequently, there were several efforts aiming at building federations of learning content and content repositories
~\cite{rehak2005model,mukherjee2005system,barcelos2011agent}. In 2005, Rehak \emph{et al.}~\cite{rehak2005model} developed a reference model describing how to establish an interoperable repository infrastructure by creating federations of repositories, where the metadata are collected from the contributing repositories into a central registry provided with a single point of discovery and access. The ultimate goal of this model is to enable learning from diverse content repositories. These practices in federated learning community or federated search service have provided effective references for the development of federated learning algorithms.

Federated learning holds great promises on healthcare data analytics. For both provider (e.g., building a model for predicting the hospital readmission risk with patient Electronic Health Records (EHR) \cite{min2019predictive}) and consumer (patient) based applications (e.g., screening atrial fibrillation with electrocardiograms captured by smartwatch \cite{perez2019large}), the sensitive patient data can stay either in local institutions or with individual consumers without going out during the federated model learning process, which effectively protects the patient privacy. The goal of this paper is to review the setup of federated learning, discuss the general solutions and challenges, as well as envision its applications in healthcare.

In this review, after a formal overview of federated learning, we summarize the main challenges and recent progress in this field. Then we illustrate the potential of federated learning methods in healthcare by describing the successful recent research. At last, we discuss the main opportunities and open questions for future applications in healthcare.

\begin{figure*}[htbp]
\centering
\includegraphics[width=1\linewidth]{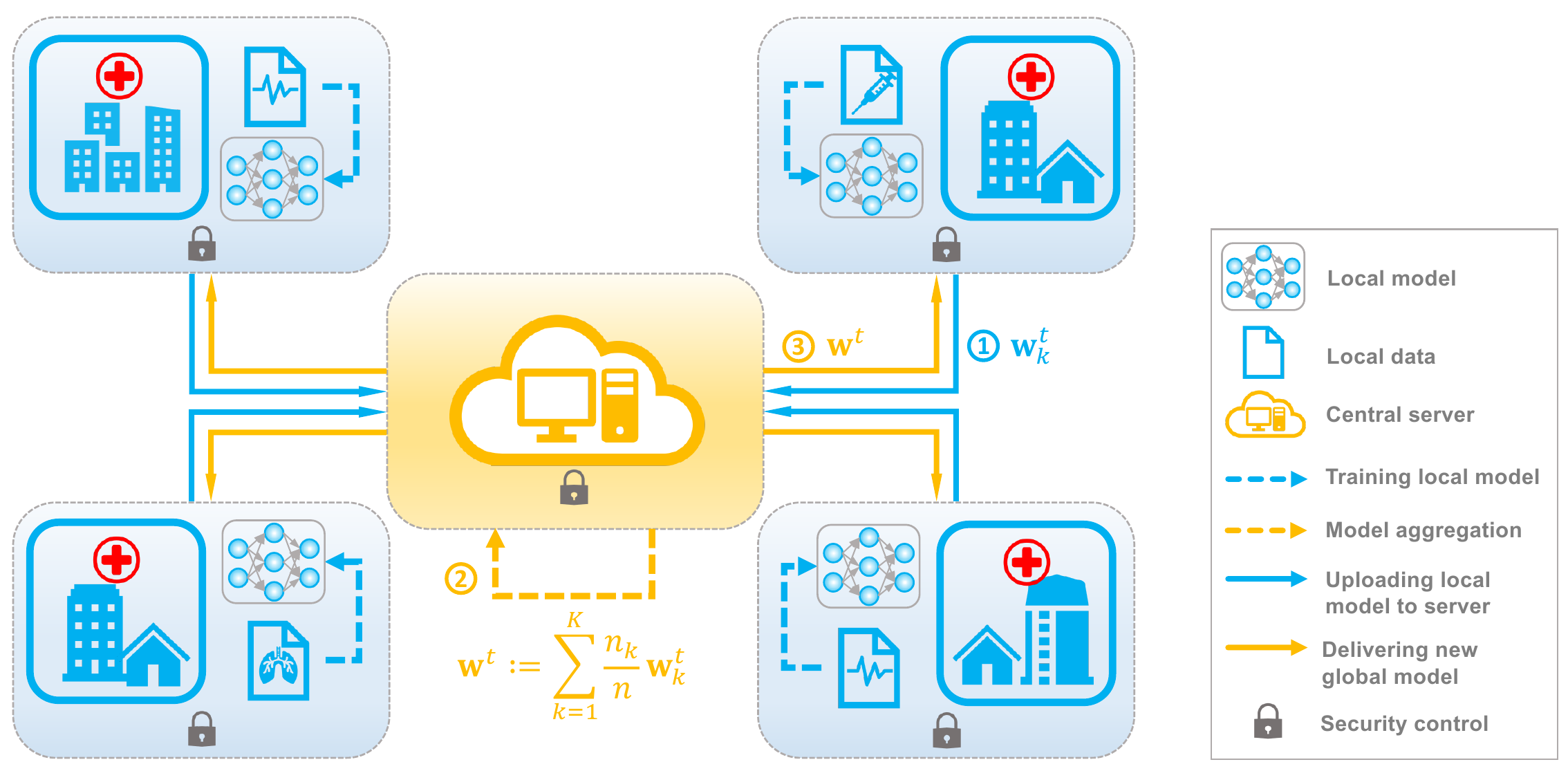}
\caption{\textbf{Schematic of the federated learning framework.} The model is trained in a distributed manner: The institutions periodically communicate the local updates with a central server to learn a global model. The central server aggregates the updates and sends back the parameters of the updated global model.}
\label{fig:fl}
\end{figure*}

\textbf{Difference with Existing Reviews} There has been a few review articles on federated learning recently. For example, Yang \emph{et al.}~\cite{Yang:2019:FML:3306498.3298981} wrote the early federated learning survey summarizing the general privacy-preserving techniques that can be applied to federated learning. Some researchers surveyed sub-problems of federated learning, \emph{e.g.,} personalization techniques~\cite{kulkarni2020survey}, semi-supervised learning algorithms~\cite{jin2020survey}, threat models~\cite{lyu2020threats}, mobile edge networks~\cite{lim2019federated}. Kairouz \emph{et al.}~\cite{kairouz2019advances} discussed recent advances and presented an extensive collection of open problems and challenges. Li~\emph{et al.}~\cite{li2019federated} conducted the review on federated learning from a system viewpoint. Different from those reviews, this paper provided the potential of federated learning to be applied in healthcare. We summarized the general solution to the challenges in federated learning scenario and surveyed a set of representative federated learning methods for healthcare. In the last part of this review, we outlined some directions or open questions in federated learning for healthcare. An early version of this paper is available on arXiv \cite{xu2019federated}.


\section{Federated Learning}
\label{sec:setting}
Federated learning is a problem of training a high-quality shared global model with a central server from decentralized data scattered among large number of different clients (Fig.~\ref{fig:fl}).
Mathematically, assume there are $K$ activated clients where the data reside in (a client could be a mobile phone, a wearable device, or a clinical institution data warehouse, \emph{etc}). Let $\mathcal{D}_k$ denote the data distribution associated to client $k$ and $n_k$ the number of samples available from that client. $n = \sum_{k=1}^K n_k$ is the total sample size. 
Federated learning problem boils down to solving a empirical risk minimization problem of the form~\cite{konevcny2015federated,konevcny2016bfederated,mcmahan2017communication}:
\begin{small}
\begin{equation}\label{eq:fl}
    \min_{\mathbf{w}\in\mathbb{R}^d} F(\mathbf{w}):=\sum_{k=1}^K \frac{n_k}{n}F_k(\mathbf{w})\ \ \ \ \text{where}\ \ \ 
    F_k(\mathbf{w}):=\frac{1}{n_k}\sum_{\mathbf{x}_i\in\mathcal{D}_k}f_i(\mathbf{w}),
\end{equation}
\end{small}
where $\mathbf{w}$ is the model parameter to be learned.

In particular, algorithms for federated learning face with a number of challenges~\cite{smith2017federated,caldas2018leaf}, specifically:
\begin{itemize}
    \item \textbf{Statistical:} The data distribution among all clients differ greatly, \emph{i.e.,} $\forall k\neq\tilde{k}$, we have $\mathbb{E}_{\mathbf{x}_i \sim {\mathcal{D}_k}}[f_i(\mathbf{w};\mathbf{x}_i)] \neq \mathbb{E}_{\mathbf{x}_i \sim \mathcal{D}_{\tilde{k}}} [f_i(\mathbf{w};\mathbf{x}_i)]$. It is such that any data points available locally are far from being a representative sample of the overall distribution, \emph{i.e.,} $\mathbb{E}_{\mathbf{x}_i \sim \mathcal{D}_k}[f_i(\mathbf{w};\mathbf{x}_i)] \neq F(\mathbf{w})$.
    \item \textbf{Communication:} The number of clients $K$ is large and can be much bigger than the average number of training sample stored in the activated clients, \emph{i.e.,} $K \gg ({n}/{K})$. 
    \item \textbf{Privacy and Security:} Additional privacy protections are needed for unreliable participating clients. It is impossible to ensure none of the millions of clients are malicious.
\end{itemize}
Next, we will survey, in detail, the existing federated learning related works on handling such challenges. 

\subsection{Statistical Challenges of Federated Learning}
\label{sec:statis}
The naive way to solve the federated learning problem is through Federated Averaging (\textit{FedAvg})~\cite{mcmahan2017communication}. It is demonstrated can work with certain non independent identical distribution (non-IID) data by requiring all the clients to share the same model.
However, \textit{FedAvg} does not address the statistical challenge of strongly skewed data distributions. The performance of convolutional neural networks trained with \textit{FedAvg} algorithm can reduce significantly due to the weight divergence~\cite{zhao2018federated}.
Existing research on dealing with the statistical challenge of federated learning can be grouped into two fields, \emph{i.e.,} consensus solution and pluralistic solution.

\subsubsection{Consensus Solution}
\label{sec:consensus}
Most centralized models are trained on the aggregated training samples obtained from the samples drawn from the local clients~\cite{smith2017federated,zhao2018federated}. Intrinsically, the centralized model is trained to minimize the loss with respect to the uniform distribution~\cite{pmlr-v97-mohri19a}: $\bar{\mathcal{D}}=\sum_{k=1}^K\frac{n_k}{n}\mathcal{D}_k$, where $\bar{\mathcal{D}}$ is the target data distribution for the learning model. However, this specific uniform distribution is not an adequate solution in most scenarios. 

To address this issue, the recent proposed solution is to model the target distribution or force the data adapt to the uniform distribution~\cite{zhao2018federated,pmlr-v97-mohri19a}. Specifically, Mohri \emph{et al.}~\cite{pmlr-v97-mohri19a} proposed a minimax optimization scheme, \emph{i.e.,} agnostic federated learning (AFL), where the centralized model is optimized for any possible target distribution formed by a mixture of the client distributions. This method has only been applied at small scales. Compared to AFL, Li \emph{et al.}~\cite{li2019fair} proposed $q$-Fair Federated Learning (\textit{q-FFL}), assigning higher weight to devices with poor performance, so that the distribution of accuracy in the network reduces in variance. They empirically demonstrate the improved flexibility and scalability of \textit{q-FFL} compared to AFL. 

Another commonly used method is globally sharing a small portion of data between all the clients~\cite{zhao2018federated,nishio2018client}. The shared subset is required containing a uniform distribution over classes from the central server to the clients. In addition to handle non-IID issue, sharing information of a small portion of trusted instances and noise patterns can guide the local agents to select compact training subset, while the clients learn to add changes to selected data samples, in order to improve the test performance of the global model~\cite{han2019robust}.

\subsubsection{Pluralistic Solution}
Generally, it is difficult to find a consensus solution $\mathbf{w}$ that is good for all components $\mathcal{D}_i$. Instead of wastefully insisting on a consensus solution, many researchers choose to embracing this heterogeneity. 

Multi-task learning (MTL) is a natural way to deal with the data drawn from different distributions. It directly captures relationships amongst non-IID and unbalanced data by leveraging the relatedness between them in comparison to learn a single global model. In order to do this, it is necessary to target a particular way in which tasks are related, \emph{e.g.} sharing sparsity, sharing low-rank structure, graph-based relatedness and so forth. Recently, Smith \emph{et al.}~\cite{smith2017federated} empirically demonstrated this point on real-world federated datasets and proposed a novel method \textit{MOCHA} to solve a general convex MTL problem with handling the system challenges at the same time. 
Later, Corinzia \emph{et al.}~\cite{corinzia2019variational} introduced \textit{VIRTUAL}, an algorithm for federated multi-task learning with non-convex models. They consider the federation of central server and clients as a Bayesian network and perform training using approximated variational inference. This work bridges the frameworks of federated and transfer/continuous learning.

The success of multi-task learning rests on whether the chosen relatedness assumptions hold. Compared to this, pluralism can be a critical tool for dealing with heterogeneous data without any additional or even low-order terms that depend on the relatedness as in MTL~\cite{eichner2019semi}. Eichner \emph{et al.}~\cite{eichner2019semi} considered training in the presence of block-cyclic data, and showed that a remarkably simple pluralistic approach can entirely resolve the source of data heterogeneity. When the component distributions are actually different, pluralism can outperform the ``ideal'' IID baseline.  


\begin{figure*}[htbp]
\centering
\includegraphics[width=0.88\linewidth]{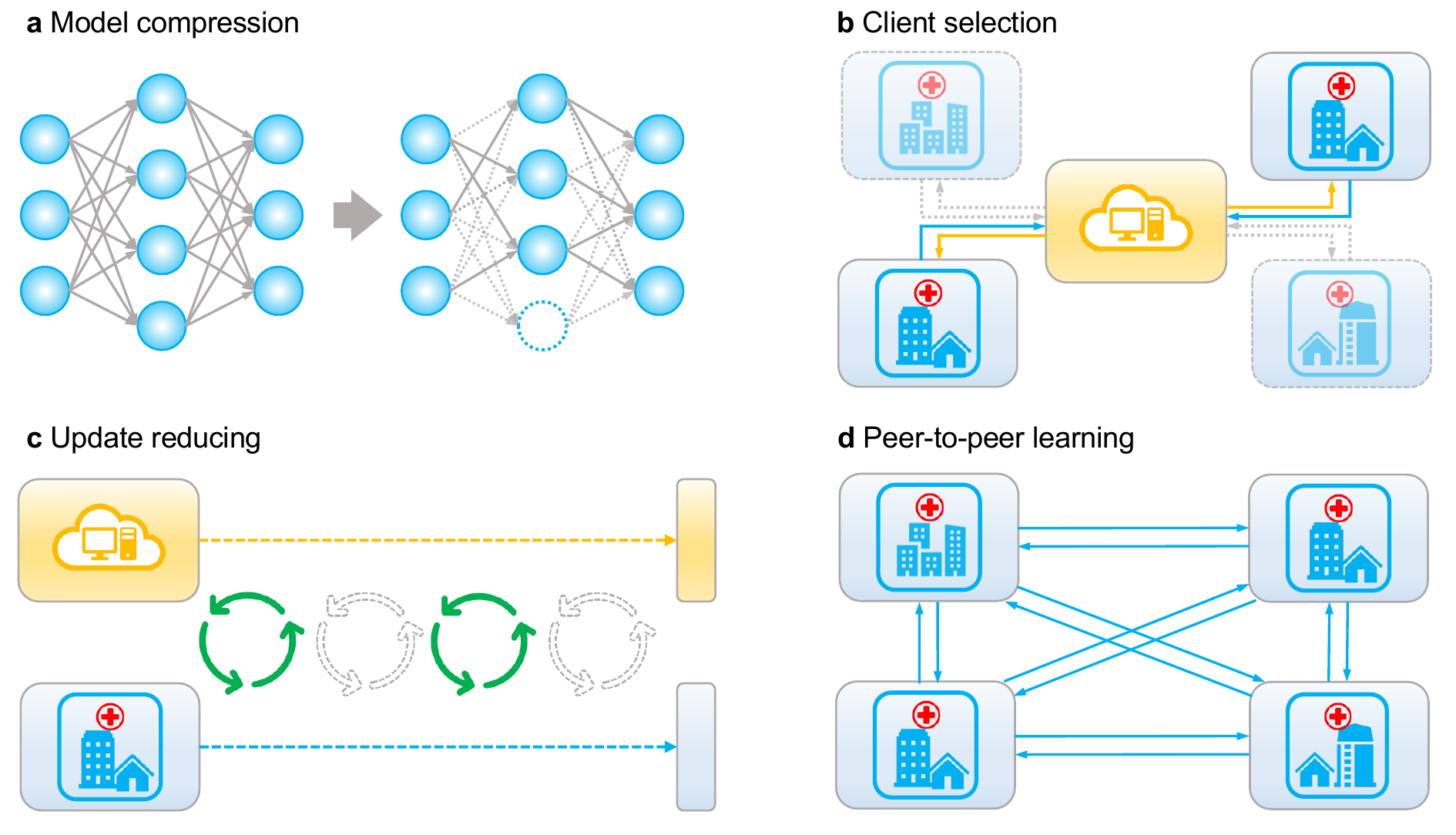}
\caption{\textbf{Communication efficient federated learning methods.} Existing research on improving communication efficiency can be categorized into \textbf{a} model compression, \textbf{b} clients selection, \textbf{c} updates reducing and \textbf{d} peer-to-peer learning.}
\label{fig:com}
\end{figure*}

\subsection{Communication Efficiency of Federated Learning}
\label{sec:communication}
In federated learning setting, training data remain distributed over a large number of clients each with unreliable and relatively slow network connections. Naively for synchronous protocol in federated learning~\cite{smith2017federated,konevcny2016afederated}, the total number of bits that required during uplink (clinets $\to$ server) and downlink (server $\to$ clients) communication by each of the $K$ clients during training are given by
\begin{equation}\label{eq:communication}
    \mathcal{B}^{up/down}\in \mathcal{O}(U \times \underbrace{|\mathbf{w}|\times (H(\triangle\mathbf{w}^{up/down})+\beta)}_{\text{update size}})
\end{equation}
where $U$ is the total number of updates performed by each client, $|\mathbf{w}|$ is the size of the model and $H(\triangle\mathbf{w}^{up/down})$ is the entropy of the weight updates exchanged during transmitting process. $\beta$ is the difference between the true update size and the minimal update size (which is given by the entropy)~\cite{sattler2019robust}. 
Apparently, we can consider three ways to reduce the communication cost: a) reduce the number of clients $K$, b) reduce the update size, c) reduce the number of updates $U$. Starting at these three points, we can organize existing research on communication-efficient federated learning into four groups, \emph{i.e.,} model compression, clients selection, updates reducing and peer-to-peer learning (Fig.~\ref{fig:com}).

\subsubsection{Client Selection.} 
The most natural and rough way for reducing communicationn cost is to restrict the participated clients or choose a fraction of parameters to be updated at each round. Shokri \emph{et al.}~\cite{shokri2015privacy} use the selective stochastic gradient descent protocol, where the selection can be completely random or only the parameters whose current values are farther away from their local optima are selected, \emph{i.e.}, those that have a larger gradient.
Nishio \emph{et al.}~\cite{nishio2018client} proposed a new protocol referred to as \textit{FedCS}, where the central server manage the resources of heterogeneous clients and determine which clients should participate the current training task by analyzing the resource information of each client, such as wireless channel states, computational capacities and the size of data resources relevant to the current task.
Here, the server should decide how much data, energy and CPU resources used by the mobile devices such that the energy consumption, training latency, and bandwidth cost are minimized while meeting requirements of the training tasks. Anh~\cite{anh2019efficient} thus propose to use the Deep Q-Learning~\cite{van2016deep} technique that enables the server to find the optimal data and energy management for the mobile devices participating in the mobile crowd-machine learning through federated learning without any prior knowledge of network dynamics. 

\subsubsection{Model Compression}
The goal of model compression is to compress the server-to-client exchanges to reduce uplink/downlink communication cost. The first way is through structured updates, where the update is directly learned from a restricted space parameterized using a smaller number of variables, \emph{e.g.} sparse, low-rank~\cite{konevcny2016afederated}, or more specifically, pruning the least useful connections in a network~\cite{han2015deep,zhu2019multi}, weight quantization~\cite{chen2019communication,sattler2019robust}, and model distillation~\cite{hinton2015distilling}.
The second way is lossy compression, where a full model update is first learned and then compressed using a combination of quantization, random rotations, and subsampling before sending it to the server~\cite{konevcny2016afederated,agarwal2018cpsgd}.
Then the server decodes the updates before doing the aggregation. 

Federated dropout, in which each client, instead of locally training an update to the whole global model, trains an update to a smaller sub-model~\cite{caldas2018expanding}. These sub-models are subsets of the global model and, as such, the computed local updates have a natural interpretation as updates to the larger global model.
Federated dropout not only reduces the downlink communication but also reduces the size of uplink updates. Moreover, the local computational costs is correspondingly reduced since the local training procedure dealing with parameters with smaller dimensions. 

\begin{figure*}[htbp]
\centering
\includegraphics[width=1\linewidth]{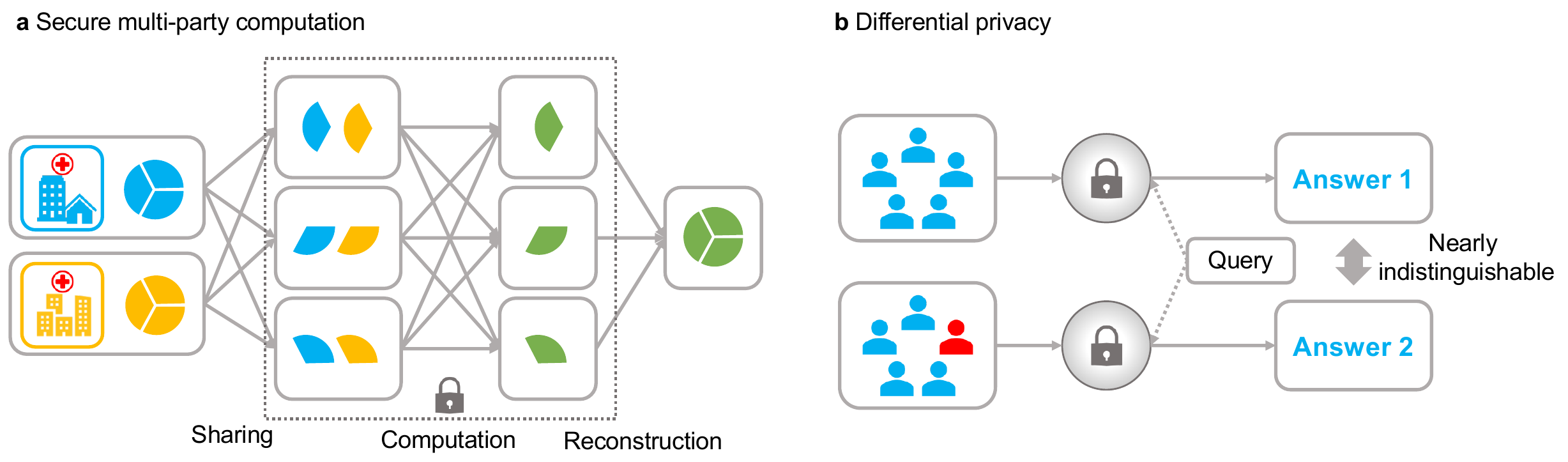}
\caption{\textbf{Privacy-preserving schemes.} \textbf{a} Secure multi-party computation. In security sharing, security values (blue and yellow pie) are split into any number of shares that are distributed among the computing nodes. During the computation, no computation node is able to recover the original value nor learn anything about the output (green pie). Any nodes can combine their shares to reconstruct the original value. \textbf{b} Differential privacy. It guarantees that anyone seeing the result of a differentially private analysis will make the same inference (Answer 1 and Answer 2 are nearly indistinguishable). }
\label{fig:privacy}
\end{figure*}

\subsubsection{Updates Reduction}
Kamp \emph{et al.}~\cite{kamp2018efficient} proposed to average models dynamically depending on the utility of the communication, which leads to a reduction of communication by an order of magnitude compared to periodically communicating state-of-the-art approaches. This facet is well suited for massively distributed systems with limited communication infrastructure.
Bui \emph{et al.}~\cite{bui2018partitioned} improved federated learning for Bayesian neural networks using partitioned variational inference, where the client can decide to upload the parameters back to the central server after multiple passes through its data, after one local epoch, or after just one mini-batch.
Guha \emph{et al.}~\cite{guha2018oneshot} focused on techniques for one-shot federated learning, in which they learn a global model from data in the network using only a single round of communication between the devices and the central server. 
Besides above works, Ren~\emph{et al.}~\cite{ren2019accelerating} theoretically analyzed the detailed expression of the learning efficiency in the CPU scenario and formulate a training acceleration problem under both communication and learning resource budget. Reinforcement learning and round robin learning are widely used to manage the communication and computation resources~\cite{anh2019efficient,wang2018edge,zhuo2019federated,ickin2019privacy}.

\subsubsection{Peer-to-Peer Learning}
In federated learning, a central server is required to coordinate the training process of the global model.
However, the communication cost to the central server may be not affordable since a large number of clients are usually involved.
Also, many practical peer-to-peer networks are usually dynamic, and it is not possible to regularly access a fixed central server. Moreover, because of the dependence on central server, all clients are required to agree on one trusted central body, and whose failure would interrupt the training process for all clients. Therefore, some researches began to study fully decentralized framework where the central server is not required~\cite{shayan2018biscotti,roy2019braintorrent,lalitha2019peer,he2019central}. The local clients are distributed over the graph/network where they only communicate with their one-hop neighbors. Each client updates its local belief based on own data, then aggregates information from the one-hop neighbors. 

\subsection{Privacy and Security}
\label{sec:privacy}
In federated learning, we usually assume the number of participated clients (\emph{e.g.}, phones, cars, clinical institutions...) is large, potentially in the thousands or millions. It is impossible to ensure none of the clients are malicious. The setting of federated learning, where the model is trained locally without revealing the input data or the model's output to any clients, prevents direct leakage while training or using the model. However, the clients may infer some information about another client's private dataset given the execution of $f(\mathbf{w})$, or over the shared predictive model $\mathbf{w}$~\cite{truex2018hybrid}. 
To this end, there have been many efforts focus on privacy either from an individual point of view or multiparty views, especially in social media field which significantly exacerbated multiparty privacy (MP) conflicts~\cite{thomas2010unfriendly,such2018multiparty}. 

\subsubsection{Secure Multi-Party Computation}
Secure multi-party computation (SMC) has a natural application to federated learning scenarios, where each individual client use a combination of cryptographic techniques and oblivious transfer to jointly compute a function of their private data~\cite{pathak2010multiparty,bonawitz2017practical}. 
Homomorphic encryption is a public key system, where any party can encrypt its data with a known public key and perform calculations with data encrypted by others with the same public key~\cite{fontaine2007survey}. Due to its success in cloud computing, it comes naturally into this realm, and it has certainly been used in many federated learning researches~\cite{hardy2017private,Chai2019Secure}.

Although SMC guarantees that none of the parties share anything with each other or with any third party, it can not prevent an adversary from learning some individual information, \emph{e.g.}, which clients' absence might change the decision boundary of a classifier, etc. Moreover, SMC protocols are usually computationally expensive even for the simplest problems, requiring iterated encryption/decryption and repeated communication between participants about some of the encrypted results~\cite{pathak2010multiparty}.

\begin{table}[htbp]
  \centering
  \caption{Summary of recent work on federated learning for healthcare}
    \begin{tabular}{lccm{3cm}}
    \hline \multicolumn{1}{l}{Problem} & \multicolumn{1}{l}{ML Method} & \multicolumn{1}{l}{\# Hospital} & \multicolumn{1}{l}{Data} \\
    \hline
        Patient Similarity Learning~\cite{lee2018privacy} &Hashing &3  &MIMIC-III~\cite{johnson2016mimic}  \\
        Patient Similarity Learning~\cite{xu2020federated} &Hashing &20  &MIMIC-III  \\
        Phenotyping~\cite{kim2017federated} &*TF &1-5  &MIMIC-III, UCSD~\cite{wah2011caltech} \\
        Phenotyping~\cite{liu2019two} &NLP &10  &MIMIC-III\\
        Representation Learning~\cite{silva2018federated} &PCA &10-100  &ADNI, UK Biobank, PPMI, MIRIAD\\
        Mortality Prediction~\cite{huang2019patient} & Autoencoder &5-50  &eICU Collaborative Research Database~\cite{pollard2018eicu}\\
        Hospitalizations Prediction~\cite{brisimi2018federated} &SVM &5, 10  &Boston Medical Center\\
        Preterm-birth  Prediction~\cite{boughorbel2019federated} &RNN &50 &Cerner Health Facts\\
        Mortality Prediction~\cite{pfohl2019federated} & LR, NN &31 &eICU Collaborative Research Database\\
        Mortality Prediction~\cite{sharma2019preserving} &LR, MLP &2 &MIMIC-III\\
    \hline
    \multicolumn{4}{l}{*TF: Tensor Factorization, MLP: Multi-layer Perceptron}
    \end{tabular}%
  \label{tab:recent_healthcare}%
\end{table}%

\subsubsection{Differential Privacy}
Differential privacy (DP)~\cite{dwork2006our} is an alternative theoretical model for protecting the privacy of individual data, which has been widely applied to many areas, not only traditional algorithms, \emph{e.g.} boosting~\cite{dwork2010boosting}, principal component analysis~\cite{chaudhuri2013near}, support vector machine~\cite{rubinstein2009learning}, but also deep learning research~\cite{abadi2016deep,mcmahan2017learning}. 
It ensures that the addition or removal does not substantially affect the outcome of any analysis, and is thus also widely studied in federated learning research to prevent the indirect leakage~\cite{shokri2015privacy,mcmahan2017learning,abadi2016deep}. However, DP only protects users from data leakage to a certain extent, and may reduce performance in prediction accuracy because it is a lossy method~\cite{cheng2019secureboost}.
Thus, some researchers combine DP with SMC to reduce the growth of noise injection as the number of parties increases without sacrificing privacy while preserving provable privacy guarantees, protecting against extraction attacks and collusion threats~\cite{cheng2019secureboost,truex2018hybrid}. 
\section{Applications}
\label{sec:app}
Federated learning has been incorporated and utilized in many domains. This widespread adoption is due in part by the fact that it enables a collaborative modeling mechanism that allows for efficient ML all while ensuring data privacy and legal compliance between multiple parties or multiple computing nodes. Some promising examples that highlight these capabilities are virtual keyboard prediction~\cite{hard2018federated,mcmahan2017learning}, smart retail~\cite{zhao2019mobile}, finance~\cite{Yang:2019:FML:3306498.3298981}, vehicle-to-vehicle communication~\cite{samarakoon2018federated}. In this section, we focus primarily on applications within the healthcare space, but also discuss promising applications in other domains since some principles can be applied to healthcare.  

\subsection{Healthcare}
EHRs have emerged as a crucial source of real world healthcare data that has been used for an amalgamation of important biomedical research~\cite{glicksberg2018next,jensen2012mining}, including for machine learning research~\cite{miotto2018deep}. While providing a huge amount of patient data for analysis, EHRs contain systemic and random biases overall and specific to hospitals that limit the generalizability of results. For example, Obermeyer \emph{et al.}~\cite{obermeyer2019dissecting} found that a commonly used algorithm to determine enrollment in specific health programs were biased against African Americans, assigning the same level of risk to healthier Caucasian patients. These improperly calibrated algorithms can arise due to a variety of reasons, such as differences in underlying access to care or low representation in training data. It is clear that one way to alleviate the risk for such biased algorithms is the ability to learn from EHR data that is more representative of the global population and which goes beyond a single hospital or site. Unfortunately, due to a myriad of reasons such as discrepant data schemes and privacy concerns, it is unlikely that data will eve be connected together in a single database to learn from all at once. The creation and utility of standardized common data models, such as OMOP~\cite{hripcsak2015observational}, allow for more wide-spread replication analyses but it does not overcome the limitations of joint data access. As such, it is imperative that alternative strategies emerge for learning from multiple EHR data sources that go beyond the common discovery-replication framework. Federated learning might be the tool to enable large-scale representative ML of EHR data and we discuss many studies which demonstrate this fact below.  

Federated learning is a viable method to connect EHR data from medical institutions, allowing them to share their experiences, and not their data, with a guarantee of privacy ~\cite{boughorbel2019federated,gruendner2019ketos,raja2014modern,duan2020learning,li2019distributed,huang2019patient}. In these scenarios, the performance of ML model will be significantly improved by the iterative improvements of learning from large and diverse medical data sets.
There have been some tasks were studied in federated learning setting in healthcare, \emph{e.g.}, patient similarity learning~\cite{lee2018privacy}, patient representation learning, phenotyping~\cite{kim2017federated,liu2019two}, predictive modeling~\cite{brisimi2018federated,huang2019patient,sharma2019preserving}, \emph{etc}.
Summary of these work is listed in Table~\ref{tab:recent_healthcare}.

\subsection{Others}
An important application of federated learning is for natural language processing (NLP) tasks. When Google first proposed federated learning concept in 2016, the application scenario is Gboard - a virtual keyboard of Google for touchscreen mobile devices with support for more than 600 language varieties~\cite{hard2018federated,mcmahan2017learning}. Indeed, as users increasingly turn to mobile devices, fast mobile input methods with auto-correction, word completion, and next-word prediction features are becoming more and more important. For these NLP tasks, especially next-word prediction, typed text in mobile apps are usually better than the data from scanned books or speech-to-text in terms of aiding typing on a mobile keyboard. However, these language data often contain sensitive information, \emph{e.g.}, passwords, search queries, or text messages with personal information. Therefore, federated learning has a promising application in NLP like virtual keyboard prediction~\cite{hard2018federated,mcmahan2017learning,bonawitz2019towards}.

\begin{table*}[htbp]
  \centering
  \caption{Popular tools for federated learning research}
    \begin{tabular}{ccm{7cm}}
    \hline \multicolumn{1}{l}{Project Name} & \multicolumn{1}{l}{Developer} & \multicolumn{1}{c}{Description}\\
    \hline
        PySyft~\cite{ryffel2018generic} & OpenMined &It decouples private data from model training using federated learning, DP and MPC within PyTorch. TensorFlow bindings is also available~\cite{pysyft2019}.\\
    \hline
        TFF~\cite{tff2019} &Google &With TFF, TensorFlow provides users with a flexible and open framework through which they can simulate distributed computing locally. \\
    \hline
        FATE~\cite{fate2019} &Webank &FATE support the Federated AI ecosystem, where a secure computing protocol is implemented based on homomorphic encryption and MPC.\\
    \hline
        Tensor/IO~\cite{tensorio2019} &Dow \emph{et al.}  &Tensor/IO is a lightweight cross-platform library for on-device machine learning, bringing the power of TensorFlow and TensorFlow Lite to iOS, Android, and React native applications.\\
    \hline
    \end{tabular}%
  \label{tab:platform}%
\end{table*}%

Other applications include smart retail~\cite{zhao2019mobile}, finance~\cite{kawa2019credit} and so on. Specifically, smart retail aims to use machine learning technology to provide personalized services to customers based on data like user purchasing power and product characteristics for product recommendation and sales services. In terms of financial applications, Tencent's WeBank 
leverages federated learning technologies for credit risk management, where several Banks could jointly generate a comprehensive credit score for a customer without sharing his or her data~\cite{Yang:2019:FML:3306498.3298981}. 
With the growth and development of federated learning, there are many companies or research teams that have carried out various tools oriented to scientific research and product development. Popular ones are listed in Table~\ref{tab:platform}.

\section{Conclusions and Open Questions}
\label{sec:conclude}
In this survey, we review the current progress on federated learning including, but not limited to healthcare field. We summarize the general solutions to the various challenges in federated learning and hope to provide a useful resource for researchers to refer. Besides the summarized general issues in federated learning setting, we list some probably encountered directions or open questions when federated learning is applied in healthcare area in the following.

\begin{itemize}
\item \textbf{Data Quality}.
Federated learning has the potential to connect all the isolated medical institutions, hospitals or devices to make them share their experiences with privacy guarantee. However, most health systems suffer from data clutter and efficiency problems. The quality of data collected from multiple sources is uneven and there is no uniform data standard. The analyzed results are apparently worthless when dirty data are accidentally used as samples. The ability to strategically leverage medical data is critical. Therefore, how to clean, correct and complete data and accordingly ensure data quality is a key to improve the machine learning model weather we are dealing with federated learning scenario or not.

\item \textbf{Incorporating Expert Knowledge}.
In 2016, IBM introduced Watson for Oncology, a tool that uses the natural language processing system to summarize patients' electronic health records and search the powerful database behind it to advise doctors on treatments. Unfortunately, some oncologists say they trust their judgment more than Watson tells them what needs to be done~\footnote{http://news.moore.ren/industry/158978.htm}. Therefore, hopefully doctors will be involved in the training process. Since every data set collected here cannot be of high quality, so it will be very helpful if the standards of evidence-based machine is introduced, doctors will also see the diagnostic criteria of artificial intelligence. If wrong, doctors will give further guidance to artificial intelligence to improve the accuracy of machine learning model during training process."

\item \textbf{Incentive Mechanisms}.
With the internet of things and the variety of third party portals, a growing number of smartphone healthcare apps are compatible with wearable devices. In addition to data accumulated in hospitals or medical centers, another type of data that is of great value is coming from wearable devices not only to the researchers, but more importantly for the owners. However, during federated model training process, the clients suffer from considerable overhead in communication and computation. 
Without well-designed incentives, self-interested mobile or other wearable devices will be reluctant to participate in federal learning tasks, which will hinder the adoption of federated learning~\cite{kang2019incentive}.  How to design an efficient incentive mechanism to attract devices with high-quality data to join federated learning is another important problem.

\item \textbf{Personalization}.
Wearable devices are more focus on public health, which means helping people who are already healthy to improve their health, such as helping them exercise, practice meditation and improve their sleep quality. How to assist patients to carry out scientifically designed personalized health management, correct the functional pathological state by examining indicators, and interrupt the pathological change process are very important. Reasonable chronic disease management can avoid emergency visits and hospitalization and reduce the number of visits. Cost and labor savings. Although there are some general work about federated learning personalization~\cite{sim2019investigation,jiang2019improving}, for healthcare informatics, how to combining the medical domain knowledge and make the global model be personalized for every medical institutions or wearable devices is another open question.

\item \textbf{Model Precision}.
Federated tries to make isolated institutions or devices share their experiences, and the performance of machine learning model will be significantly improved by the formed large medical dataset. However, the prediction task is currently restricted and relatively simple. Medical treatment itself is a very professional and accurate field. Medical devices in hospitals have incomparable advantages over wearable devices. And the models of Doc.ai could predict the phenome collection of one's biometric data based on its selfie, such as height, weight, age, sex and BMI\footnote{https://doc.ai/blog/do-you-know-how-valuable-your-medical-da/}. How to improve the prediction model to predict future health conditions is definitely worth exploring.
\end{itemize}

\begin{acknowledgements}
The work is supported by ONR N00014-18-1-2585 and NSF 1750326.
\end{acknowledgements}

%
\section*{Conflict of interest}
The authors declare that they have no conflict of interest.

\bibliographystyle{spmpsci}      
\bibliography{ref}

\begin{thebibliography}{100}
\providecommand{\url}[1]{{#1}}
\providecommand{\urlprefix}{URL }
\expandafter\ifx\csname urlstyle\endcsname\relax
  \providecommand{\doi}[1]{DOI~\discretionary{}{}{}#1}\else
  \providecommand{\doi}{DOI~\discretionary{}{}{}\begingroup
  \urlstyle{rm}\Url}\fi

\bibitem{abadi2016deep}
Abadi, M., Chu, A., Goodfellow, I., McMahan, H.B., Mironov, I., Talwar, K.,
  Zhang, L.: Deep learning with differential privacy.
\newblock In: Proceedings of the 2016 ACM SIGSAC Conference on Computer and
  Communications Security, pp. 308--318. ACM (2016)

\bibitem{agarwal2018cpsgd}
Agarwal, N., Suresh, A.T., Yu, F.X.X., Kumar, S., McMahan, B.: cpsgd:
  Communication-efficient and differentially-private distributed sgd.
\newblock In: Advances in Neural Information Processing Systems, pp. 7564--7575
  (2018)

\bibitem{fate2019}
AI, W.: Federated ai technology enabler.
\newblock \url{https://www.fedai.org/cn/} (2019)

\bibitem{anh2019efficient}
Anh, T.T., Luong, N.C., Niyato, D., Kim, D.I., Wang, L.C.: Efficient training
  management for mobile crowd-machine learning: A deep reinforcement learning
  approach.
\newblock IEEE Wireless Communications Letters  (2019)

\bibitem{barcelos2011agent}
Barcelos, C., Gluz, J., Vicari, R.: An agent-based federated learning object
  search service.
\newblock Interdisciplinary journal of e-learning and learning objects
  \textbf{7}(1), 37--54 (2011)

\bibitem{bonawitz2019towards}
Bonawitz, K., Eichner, H., Grieskamp, W., Huba, D., Ingerman, A., Ivanov, V.,
  Kiddon, C., Konecny, J., Mazzocchi, S., McMahan, H.B., et~al.: Towards
  federated learning at scale: System design.
\newblock arXiv preprint arXiv:1902.01046  (2019)

\bibitem{bonawitz2017practical}
Bonawitz, K., Ivanov, V., Kreuter, B., Marcedone, A., McMahan, H.B., Patel, S.,
  Ramage, D., Segal, A., Seth, K.: Practical secure aggregation for
  privacy-preserving machine learning.
\newblock In: Proceedings of the 2017 ACM SIGSAC Conference on Computer and
  Communications Security, pp. 1175--1191. ACM (2017)

\bibitem{boughorbel2019federated}
Boughorbel, S., Jarray, F., Venugopal, N., Moosa, S., Elhadi, H., Makhlouf, M.:
  Federated uncertainty-aware learning for distributed hospital ehr data.
\newblock arXiv preprint arXiv:1910.12191  (2019)

\bibitem{brisimi2018federated}
Brisimi, T.S., Chen, R., Mela, T., Olshevsky, A., Paschalidis, I.C., Shi, W.:
  Federated learning of predictive models from federated electronic health
  records.
\newblock International journal of medical informatics \textbf{112}, 59--67
  (2018)

\bibitem{bui2018partitioned}
Bui, T.D., Nguyen, C.V., Swaroop, S., Turner, R.E.: Partitioned variational
  inference: A unified framework encompassing federated and continual learning.
\newblock arXiv preprint arXiv:1811.11206  (2018)

\bibitem{caldas2018expanding}
Caldas, S., Kone{\v{c}}ny, J., McMahan, H.B., Talwalkar, A.: Expanding the
  reach of federated learning by reducing client resource requirements.
\newblock arXiv preprint arXiv:1812.07210  (2018)

\bibitem{caldas2018leaf}
Caldas, S., Wu, P., Li, T., Kone{\v{c}}n{\`y}, J., McMahan, H.B., Smith, V.,
  Talwalkar, A.: Leaf: A benchmark for federated settings.
\newblock arXiv preprint arXiv:1812.01097  (2018)

\bibitem{Chai2019Secure}
Chai, D., Wang, L., Chen, K., Yang, Q.: Secure federated matrix factorization
  (2019)

\bibitem{chaudhuri2013near}
Chaudhuri, K., Sarwate, A.D., Sinha, K.: A near-optimal algorithm for
  differentially-private principal components.
\newblock The Journal of Machine Learning Research \textbf{14}(1), 2905--2943
  (2013)

\bibitem{chen2019communication}
Chen, Y., Sun, X., Jin, Y.: Communication-efficient federated deep learning
  with asynchronous model update and temporally weighted aggregation.
\newblock arXiv preprint arXiv:1903.07424  (2019)

\bibitem{cheng2019secureboost}
Cheng, K., Fan, T., Jin, Y., Liu, Y., Chen, T., Yang, Q.: Secureboost: A
  lossless federated learning framework.
\newblock arXiv preprint arXiv:1901.08755  (2019)

\bibitem{corinzia2019variational}
Corinzia, L., Buhmann, J.M.: Variational federated multi-task learning.
\newblock arXiv preprint arXiv:1906.06268  (2019)

\bibitem{tensorio2019}
doc.ai: Declarative, on-device machine learning for ios, android, and react
  native.
\newblock \url{https://github.com/doc-ai/tensorio} (2019)

\bibitem{duan2020learning}
Duan, R., Boland, M.R., Liu, Z., Liu, Y., Chang, H.H., Xu, H., Chu, H., Schmid,
  C.H., Forrest, C.B., Holmes, J.H., et~al.: Learning from electronic health
  records across multiple sites: A communication-efficient and
  privacy-preserving distributed algorithm.
\newblock Journal of the American Medical Informatics Association
  \textbf{27}(3), 376--385 (2020)

\bibitem{dwork2006our}
Dwork, C., Kenthapadi, K., McSherry, F., Mironov, I., Naor, M.: Our data,
  ourselves: Privacy via distributed noise generation.
\newblock In: Annual International Conference on the Theory and Applications of
  Cryptographic Techniques, pp. 486--503. Springer (2006)

\bibitem{dwork2010boosting}
Dwork, C., Rothblum, G.N., Vadhan, S.: Boosting and differential privacy.
\newblock In: 2010 IEEE 51st Annual Symposium on Foundations of Computer
  Science, pp. 51--60. IEEE (2010)

\bibitem{eichner2019semi}
Eichner, H., Koren, T., McMahan, H.B., Srebro, N., Talwar, K.: Semi-cyclic
  stochastic gradient descent.
\newblock arXiv preprint arXiv:1904.10120  (2019)

\bibitem{fontaine2007survey}
Fontaine, C., Galand, F.: A survey of homomorphic encryption for
  nonspecialists.
\newblock EURASIP Journal on Information Security \textbf{2007}, 15 (2007)

\bibitem{glicksberg2018next}
Glicksberg, B.S., Johnson, K.W., Dudley, J.T.: The next generation of precision
  medicine: observational studies, electronic health records, biobanks and
  continuous monitoring.
\newblock Human molecular genetics \textbf{27}(R1), R56--R62 (2018)

\bibitem{tff2019}
Google: Tensorflow federated.
\newblock \url{https://www.tensorflow.org/federated} (2019)

\bibitem{gostin2001national}
Gostin, L.O.: National health information privacy: regulations under the health
  insurance portability and accountability act.
\newblock {JAMA} \textbf{285}(23), 3015--3021 (2001)

\bibitem{gruendner2019ketos}
Gruendner, J., Schwachhofer, T., Sippl, P., Wolf, N., Erpenbeck, M., Gulden,
  C., Kapsner, L.A., Zierk, J., Mate, S., St{\"u}rzl, M., et~al.: Ketos:
  Clinical decision support and machine learning as a service--a training and
  deployment platform based on docker, omop-cdm, and fhir web services.
\newblock PloS one \textbf{14}(10) (2019)

\bibitem{guha2018oneshot}
Guha, N., Talwalkar, A., Smith, V.: One-shot federated learning.
\newblock arXiv preprint arXiv:1902.11175  (2019)

\bibitem{han2015deep}
Han, S., Mao, H., Dally, W.J.: Deep compression: Compressing deep neural
  networks with pruning, trained quantization and huffman coding.
\newblock arXiv preprint arXiv:1510.00149  (2015)

\bibitem{han2019robust}
Han, Y., Zhang, X.: Robust federated training via collaborative machine
  teaching using trusted instances.
\newblock arXiv preprint arXiv:1905.02941  (2019)

\bibitem{hard2018federated}
Hard, A., Rao, K., Mathews, R., Beaufays, F., Augenstein, S., Eichner, H.,
  Kiddon, C., Ramage, D.: Federated learning for mobile keyboard prediction.
\newblock arXiv preprint arXiv:1811.03604  (2018)

\bibitem{hardy2017private}
Hardy, S., Henecka, W., Ivey-Law, H., Nock, R., Patrini, G., Smith, G., Thorne,
  B.: Private federated learning on vertically partitioned data via entity
  resolution and additively homomorphic encryption.
\newblock arXiv preprint arXiv:1711.10677  (2017)

\bibitem{he2019central}
He, C., Tan, C., Tang, H., Qiu, S., Liu, J.: Central server free federated
  learning over single-sided trust social networks.
\newblock arXiv preprint arXiv:1910.04956  (2019)

\bibitem{hill1985rationale}
Hill, P.: The rationale for learning communities and learning community models.
   (1985)

\bibitem{hinton2015distilling}
Hinton, G., Vinyals, O., Dean, J.: Distilling the knowledge in a neural
  network.
\newblock arXiv preprint arXiv:1503.02531  (2015)

\bibitem{hripcsak2015observational}
Hripcsak, G., Duke, J.D., Shah, N.H., Reich, C.G., Huser, V., Schuemie, M.J.,
  Suchard, M.A., Park, R.W., Wong, I.C.K., Rijnbeek, P.R., et~al.:
  Observational health data sciences and informatics (ohdsi): opportunities for
  observational researchers.
\newblock Studies in health technology and informatics \textbf{216}, 574 (2015)

\bibitem{huang2019patient}
Huang, L., Liu, D.: Patient clustering improves efficiency of federated machine
  learning to predict mortality and hospital stay time using distributed
  electronic medical records.
\newblock arXiv preprint arXiv:1903.09296  (2019)

\bibitem{ickin2019privacy}
Ickin, S., Vandikas, K., Fiedler, M.: Privacy preserving qoe modeling using
  collaborative learning.
\newblock arXiv preprint arXiv:1906.09248  (2019)

\bibitem{jensen2012mining}
Jensen, P.B., Jensen, L.J., Brunak, S.: Mining electronic health records:
  towards better research applications and clinical care.
\newblock Nature Reviews Genetics \textbf{13}(6), 395--405 (2012)

\bibitem{jiang2019improving}
Jiang, Y., Kone{\v{c}}n{\`y}, J., Rush, K., Kannan, S.: Improving federated
  learning personalization via model agnostic meta learning.
\newblock arXiv preprint arXiv:1909.12488v1  (2019)

\bibitem{jin2020survey}
Jin, Y., Wei, X., Liu, Y., Yang, Q.: A survey towards federated semi-supervised
  learning.
\newblock arXiv preprint arXiv:2002.11545  (2020)

\bibitem{johnson2016mimic}
Johnson, A.E., Pollard, T.J., Shen, L., Li-wei, H.L., Feng, M., Ghassemi, M.,
  Moody, B., Szolovits, P., Celi, L.A., Mark, R.G.: Mimic-iii, a freely
  accessible critical care database.
\newblock Scientific data \textbf{3}, 160035 (2016)

\bibitem{kairouz2019advances}
Kairouz, P., McMahan, H.B., Avent, B., Bellet, A., Bennis, M., Bhagoji, A.N.,
  Bonawitz, K., Charles, Z., Cormode, G., Cummings, R., et~al.: Advances and
  open problems in federated learning.
\newblock arXiv preprint arXiv:1912.04977  (2019)

\bibitem{kamp2018efficient}
Kamp, M., Adilova, L., Sicking, J., H{\"u}ger, F., Schlicht, P., Wirtz, T.,
  Wrobel, S.: Efficient decentralized deep learning by dynamic model averaging.
\newblock In: Joint European Conference on Machine Learning and Knowledge
  Discovery in Databases, pp. 393--409. Springer (2018)

\bibitem{kang2019incentive}
Kang, J., Xiong, Z., Niyato, D., Yu, H., Liang, Y.C., Kim, D.I.: Incentive
  design for efficient federated learning in mobile networks: A contract theory
  approach.
\newblock arXiv preprint arXiv:1905.07479  (2019)

\bibitem{kawa2019credit}
Kawa, D., Punyani, S., Nayak, P., Karkera, A., Jyotinagar, V.: Credit risk
  assessment from combined bank records using federated learning  (2019)

\bibitem{kim2017federated}
Kim, Y., Sun, J., Yu, H., Jiang, X.: Federated tensor factorization for
  computational phenotyping.
\newblock In: Proceedings of the 23rd ACM SIGKDD International Conference on
  Knowledge Discovery and Data Mining, pp. 887--895. ACM (2017)

\bibitem{konevcny2015federated}
Kone{\v{c}}n{\`y}, J., McMahan, B., Ramage, D.: Federated optimization:
  Distributed optimization beyond the datacenter.
\newblock arXiv preprint arXiv:1511.03575  (2015)

\bibitem{konevcny2016bfederated}
Kone{\v{c}}n{\`y}, J., McMahan, H.B., Ramage, D., Richt{\'a}rik, P.: Federated
  optimization: Distributed machine learning for on-device intelligence.
\newblock arXiv preprint arXiv:1610.02527  (2016)

\bibitem{konevcny2016afederated}
Kone{\v{c}}n{\`y}, J., McMahan, H.B., Yu, F.X., Richt{\'a}rik, P., Suresh,
  A.T., Bacon, D.: Federated learning: Strategies for improving communication
  efficiency.
\newblock arXiv preprint arXiv:1610.05492  (2016)

\bibitem{kulkarni2020survey}
Kulkarni, V., Kulkarni, M., Pant, A.: Survey of personalization techniques for
  federated learning.
\newblock arXiv preprint arXiv:2003.08673  (2020)

\bibitem{lalitha2019peer}
Lalitha, A., Kilinc, O.C., Javidi, T., Koushanfar, F.: Peer-to-peer federated
  learning on graphs.
\newblock rXiv preprint arXiv:1901.11173  (2019)

\bibitem{lecun2015deep}
LeCun, Y., Bengio, Y., Hinton, G.: Deep learning.
\newblock nature \textbf{521}(7553), 436--444 (2015)

\bibitem{lee2018privacy}
Lee, J., Sun, J., Wang, F., Wang, S., Jun, C.H., Jiang, X.: Privacy-preserving
  patient similarity learning in a federated environment: development and
  analysis.
\newblock JMIR medical informatics \textbf{6}(2), e20 (2018)

\bibitem{li2019federated}
Li, T., Sahu, A.K., Zaheer, M., Sanjabi, M., Talwalkar, A., Smith1, V.:
  Federated optimization for heterogeneous networks.
\newblock arXiv preprint arXiv:1812.06127  (2019)

\bibitem{li2019fair}
Li, T., Sanjabi, M., Smith, V.: Fair resource allocation in federated learning.
\newblock arXiv preprint arXiv:1905.10497  (2019)

\bibitem{li2019distributed}
Li, Z., Roberts, K., Jiang, X., Long, Q.: Distributed learning from multiple
  ehr databases: Contextual embedding models for medical events.
\newblock Journal of biomedical informatics \textbf{92}, 103138 (2019)

\bibitem{lim2019federated}
Lim, W.Y.B., Luong, N.C., Hoang, D.T., Jiao, Y., Liang, Y.C., Yang, Q., Niyato,
  D., Miao, C.: Federated learning in mobile edge networks: A comprehensive
  survey.
\newblock arXiv preprint arXiv:1909.11875  (2019)

\bibitem{liu2019two}
Liu, D., Dligach, D., Miller, T.: Two-stage federated phenotyping and patient
  representation learning.
\newblock arXiv preprint arXiv:1908.05596  (2019)

\bibitem{lyu2020threats}
Lyu, L., Yu, H., Yang, Q.: Threats to federated learning: A survey.
\newblock arXiv preprint arXiv:2003.02133  (2020)

\bibitem{mcmahan2017communication}
McMahan, B., Moore, E., Ramage, D., Hampson, S., y~Arcas, B.A.:
  Communication-efficient learning of deep networks from decentralized data.
\newblock In: Artificial Intelligence and Statistics, pp. 1273--1282 (2017)

\bibitem{mcmahan2017learning}
McMahan, H.B., Ramage, D., Talwar, K., Zhang, L.: Learning differentially
  private recurrent language models.
\newblock arXiv preprint arXiv:1710.06963  (2017)

\bibitem{min2019predictive}
Min, X., Yu, B., Wang, F.: Predictive modeling of the hospital readmission risk
  from patients’ claims data using machine learning: A case study on copd.
\newblock Scientific reports \textbf{9}(1), 2362 (2019)

\bibitem{miotto2018deep}
Miotto, R., Wang, F., Wang, S., Jiang, X., Dudley, J.T.: Deep learning for
  healthcare: review, opportunities and challenges.
\newblock Briefings in bioinformatics \textbf{19}(6), 1236--1246 (2018)

\bibitem{pmlr-v97-mohri19a}
Mohri, M., Sivek, G., Suresh, A.T.: Agnostic federated learning.
\newblock In: K.~Chaudhuri, R.~Salakhutdinov (eds.) Proceedings of the 36th
  International Conference on Machine Learning, \emph{Proceedings of Machine
  Learning Research}, vol.~97, pp. 4615--4625. PMLR, Long Beach, California,
  USA (2019)

\bibitem{mukherjee2005system}
Mukherjee, R., Jaffe, H.: System and method for dynamic context-sensitive
  federated search of multiple information repositories (2005).
\newblock US Patent App. 10/743,196

\bibitem{nishio2018client}
Nishio, T., Yonetani, R.: Client selection for federated learning with
  heterogeneous resources in mobile edge.
\newblock arXiv preprint arXiv:1804.08333  (2018)

\bibitem{obermeyer2019dissecting}
Obermeyer, Z., Powers, B., Vogeli, C., Mullainathan, S.: Dissecting racial bias
  in an algorithm used to manage the health of populations.
\newblock Science \textbf{366}(6464), 447--453 (2019)

\bibitem{pysyft2019}
OpenMined: Pysyft-tensorflow.
\newblock \url{https://github.com/OpenMined/PySyft-TensorFlow} (2019)

\bibitem{pathak2010multiparty}
Pathak, M., Rane, S., Raj, B.: Multiparty differential privacy via aggregation
  of locally trained classifiers.
\newblock In: Advances in Neural Information Processing Systems, pp. 1876--1884
  (2010)

\bibitem{perez2019large}
Perez, M.V., Mahaffey, K.W., Hedlin, H., Rumsfeld, J.S., Garcia, A., Ferris,
  T., Balasubramanian, V., Russo, A.M., Rajmane, A., Cheung, L., et~al.:
  Large-scale assessment of a smartwatch to identify atrial fibrillation.
\newblock New England Journal of Medicine \textbf{381}(20), 1909--1917 (2019)

\bibitem{pfohl2019federated}
Pfohl, S.R., Dai, A.M., Heller, K.: Federated and differentially private
  learning for electronic health records.
\newblock arXiv preprint arXiv:1911.05861  (2019)

\bibitem{pollard2018eicu}
Pollard, T.J., Johnson, A.E., Raffa, J.D., Celi, L.A., Mark, R.G., Badawi, O.:
  The eicu collaborative research database, a freely available multi-center
  database for critical care research.
\newblock Scientific data \textbf{5}, 180178 (2018)

\bibitem{raja2014modern}
Raja, P.V., Sivasankar, E.: Modern framework for distributed healthcare data
  analytics based on hadoop.
\newblock In: Information and Communication Technology-EurAsia Conference, pp.
  348--355. Springer (2014)

\bibitem{rehak2005model}
Rehak, D., Dodds, P., Lannom, L.: A model and infrastructure for federated
  learning content repositories.
\newblock In: Interoperability of Web-Based Educational Systems Workshop, vol.
  143. Citeseer (2005)

\bibitem{ren2019accelerating}
Ren, J., Yu, G., Ding, G.: Accelerating dnn training in wireless federated edge
  learning system.
\newblock arXiv preprint arXiv:1905.09712  (2019)

\bibitem{roy2019braintorrent}
Roy, A.G., Siddiqui, S., P{\"o}lsterl, S., Navab, N., Wachinger, C.:
  Braintorrent: A peer-to-peer environment for decentralized federated
  learning.
\newblock arXiv preprint arXiv:1905.06731  (2019)

\bibitem{rubinstein2009learning}
Rubinstein, B.I., Bartlett, P.L., Huang, L., Taft, N.: Learning in a large
  function space: Privacy-preserving mechanisms for svm learning.
\newblock arXiv preprint arXiv:0911.5708  (2009)

\bibitem{ryffel2018generic}
Ryffel, T., Trask, A., Dahl, M., Wagner, B., Mancuso, J., Rueckert, D.,
  Passerat-Palmbach, J.: A generic framework for privacy preserving deep
  learning.
\newblock arXiv preprint arXiv:1811.04017  (2018)

\bibitem{samarakoon2018federated}
Samarakoon, S., Bennis, M., Saad, W., Debbah, M.: Federated learning for
  ultra-reliable low-latency v2v communications.
\newblock In: 2018 IEEE Global Communications Conference (GLOBECOM), pp. 1--7.
  IEEE (2018)

\bibitem{sattler2019robust}
Sattler, F., Wiedemann, S., M{\"u}ller, K.R., Samek, W.: Robust and
  communication-efficient federated learning from non-iid data.
\newblock arXiv preprint arXiv:1903.02891  (2019)

\bibitem{sharma2019preserving}
Sharma, P., Shamout, F.E., Clifton, D.A.: Preserving patient privacy while
  training a predictive model of in-hospital mortality.
\newblock arXiv preprint arXiv:1912.00354  (2019)

\bibitem{shayan2018biscotti}
Shayan, M., Fung, C., Yoon, C.J., Beschastnikh, I.: Biscotti: A ledger for
  private and secure peer-to-peer machine learning.
\newblock arXiv preprint arXiv:1811.09904  (2018)

\bibitem{shokri2015privacy}
Shokri, R., Shmatikov, V.: Privacy-preserving deep learning.
\newblock In: Proceedings of the 22nd ACM SIGSAC conference on computer and
  communications security, pp. 1310--1321. ACM (2015)

\bibitem{silva2018federated}
Silva, S., Gutman, B., Romero, E., Thompson, P.M., Altmann, A., Lorenzi, M.:
  Federated learning in distributed medical databases: Meta-analysis of
  large-scale subcortical brain data.
\newblock arXiv preprint arXiv:1810.08553  (2018)

\bibitem{sim2019investigation}
Sim, K.C., Zadrazil, P., Beaufays, F.: An investigation into on-device
  personalization of end-to-end automatic speech recognition models.
\newblock arXiv preprint arXiv:1909.06678  (2019)

\bibitem{smith2017federated}
Smith, V., Chiang, C.K., Sanjabi, M., Talwalkar, A.S.: Federated multi-task
  learning.
\newblock In: Advances in Neural Information Processing Systems, pp. 4424--4434
  (2017)

\bibitem{such2018multiparty}
Such, J.M., Criado, N.: Multiparty privacy in social media.
\newblock Commun. ACM \textbf{61}(8), 74--81 (2018)

\bibitem{thomas2010unfriendly}
Thomas, K., Grier, C., Nicol, D.M.: unfriendly: Multi-party privacy risks in
  social networks.
\newblock In: International Symposium on Privacy Enhancing Technologies
  Symposium, pp. 236--252. Springer (2010)

\bibitem{truex2018hybrid}
Truex, S., Baracaldo, N., Anwar, A., Steinke, T., Ludwig, H., Zhang, R.: A
  hybrid approach to privacy-preserving federated learning.
\newblock arXiv preprint arXiv:1812.03224  (2018)

\bibitem{van2016deep}
Van~Hasselt, H., Guez, A., Silver, D.: Deep reinforcement learning with double
  q-learning.
\newblock In: Thirtieth AAAI conference on artificial intelligence (2016)

\bibitem{wah2011caltech}
Wah, C., Branson, S., Welinder, P., Perona, P., Belongie, S.: The caltech-ucsd
  birds-200-2011 dataset  (2011)

\bibitem{wang2019ai}
Wang, F., Preininger, A.: Ai in health: State of the art, challenges, and
  future directions.
\newblock Yearbook of medical informatics \textbf{28}(01), 016--026 (2019)

\bibitem{wang2018edge}
Wang, X., Han, Y., Wang, C., Zhao, Q., Chen, X., Chen, M.: In-edge ai:
  Intelligentizing mobile edge computing, caching and communication by
  federated learning.
\newblock arXiv preprint arXiv:1809.07857  (2018)

\bibitem{xu2019federated}
Xu, J., Wang, F.: Federated learning for healthcare informatics.
\newblock arXiv preprint arXiv:1911.06270  (2019)

\bibitem{xu2020federated}
Xu, J., Xu, Z., Walker, P., Wang, F.: Federated patient hashing.
\newblock In: AAAI, pp. 6486--6493 (2020)

\bibitem{Yang:2019:FML:3306498.3298981}
Yang, Q., Liu, Y., Chen, T., Tong, Y.: Federated machine learning: Concept and
  applications.
\newblock ACM Trans. Intell. Syst. Technol. \textbf{10}(2), 12:1--12:19 (2019).
\newblock \doi{10.1145/3298981}.
\newblock \urlprefix\url{http://doi.acm.org/10.1145/3298981}

\bibitem{zhao2018federated}
Zhao, Y., Li, M., Lai, L., Suda, N., Civin, D., Chandra, V.: Federated learning
  with non-iid data.
\newblock arXiv preprint arXiv:1806.00582  (2018)

\bibitem{zhao2019mobile}
Zhao, Y., Zhao, J., Jiang, L., Tan, R., Niyato, D.: Mobile edge computing,
  blockchain and reputation-based crowdsourcing iot federated learning: A
  secure, decentralized and privacy-preserving system.
\newblock arXiv preprint arXiv:1906.10893  (2019)

\bibitem{zhu2019multi}
Zhu, H., Jin, Y.: Multi-objective evolutionary federated learning.
\newblock IEEE transactions on neural networks and learning systems  (2019)

\bibitem{zhuo2019federated}
Zhuo, H.H., Feng, W., Xu, Q., Yang, Q., Lin, Y.: Federated reinforcement
  learning.
\newblock rXiv preprint arXiv:1901.08277  (2019)

\end{thebibliography}

\end{document}